\documentclass[11pt]{article}
\usepackage{graphicx}
\usepackage{amsmath,amssymb}
\usepackage{multirow}
\usepackage{booktabs}
\usepackage{placeins}
\begin{document}

\title{Motion- and Gravity-Aware Feedforward 3D Reconstruction}

\author{
Shiyao Qian\thanks{Huawei Noah's Ark Lab, Markham, ON, Canada. Email: \texttt{\{yuan.ren3, baidongfeng\}@huawei.com}.}\hspace{0.35em}\thanks{University of Toronto, Toronto, ON, Canada. This work was done during an internship at Huawei Noah's Ark Lab. Email: \texttt{shiyao.qian@mail.utoronto.ca}.}\hspace{0.35em},
Yuan Ren\footnotemark[1]\hspace{0.35em}, Dongfeng Bai\footnotemark[1]
}
\date{}

\maketitle

\begin{abstract}
Feed-forward transformer models have recently achieved strong performance in multi-view 3D reconstruction, but they still rely primarily on appearance-based attention and often struggle in dynamic scenes with weak geometric cues. In this work, we introduce a motion- and gravity-aware framework under the Learning Using Privileged Information (LUPI) paradigm. During training, residual optical flow and gravity direction are provided as privileged signals to guide geometric representation learning; during inference, the model uses standard visual inputs only.

We add these priors to the transformer backbones with few architectural changes while preserving feed-forward efficiency. Experiments on standard benchmarks show consistent gains in reconstruction accuracy and robustness, especially in scenes with dynamic objects, ambiguous structure, and challenging camera motion. Our analysis further shows that the optimal injection stage is backbone-dependent: middle-layer integration is generally preferable for VGGT, while late-stage integration is more effective for $\pi^3$. These results suggest that LUPI offers a practical way to inject motion and physical priors into feed-forward 3D reconstruction without increasing test-time complexity.
\end{abstract}


\section{Introduction}
Feedforward transformer-based approaches have recently achieved impressive progress in multi-view 3D reconstruction by directly reasoning over dense patch correspondences across images. Methods such as VGGT\cite{wang2025vggt} demonstrate that strong geometric priors can emerge from large-scale training, enabling accurate reconstruction without computationally expensive Bundle Adjustment \cite{triggs1999bundle} involved in Structure-from-Motion \cite{schoenberger2016sfm} and dense matching with cost volumes in the Multi-view stereo methods \cite{schoenberger2016mvs}. This paradigm has significantly improved efficiency and scalability, making feedforward reconstruction an increasingly attractive solution for real-world perception tasks.

Despite these advances, most existing feedforward models rely primarily on appearance-driven attention mechanisms and treat each input view as a collection of static visual tokens. In practice, however, real-world scenes are rarely static: independently moving objects, camera ego-motion, and temporal inconsistencies introduce challenges that cannot always be resolved from appearance alone. Without explicit motion reasoning, the model must implicitly infer whether observed changes arise from viewpoint variation or true scene dynamics, which can lead to unstable correspondences and degraded reconstruction quality in dynamic or ambiguous scenarios.

In addition, feedforward reconstruction methods typically operate without global physical orientation cues. As a result, the predicted geometry may suffer from scale ambiguity, drifting orientation, or inconsistent spatial layouts across views, especially when visual evidence is weak or repetitive. Incorporating simple physical priors, such as gravity direction, offers a natural way to stabilize spatial reasoning and enforce a consistent reference frame, complementing purely appearance-based token interactions.

Motivated by the importance of motion and orientation cues in robust visual understanding, we propose to augment 3D vision foundation models with explicit motion- and gravity-aware signals. Specifically, we inject optical flow as an additional motion representation that highlights local pixel displacement patterns, enabling the model to better separate static background structure from dynamic foreground regions. Furthermore, we incorporate gravity direction as a lightweight global orientation prior, providing a stable reference axis that improves spatial consistency across views. These cues introduce structured inductive biases that are absent in standard feedforward architectures, while preserving the simplicity and efficiency of transformer-based reconstruction.

To maintain architectural integrity and isolate the impact of our priors, we adopt the Learning Using Privileged Information (LUPI) paradigm during fine-tuning and treat the new modalities as privileged data. We demonstrate generalizability by applying this approach to two architectures with distinct structural designs. Comprehensive evaluations on standard benchmarks reveal consistent gains in reconstruction accuracy and robustness, especially in scenarios with dynamic content, complex camera motion, and scale ambiguity. These results indicate that explicitly incorporating motion and physical orientation cues is a robust and generalizable strategy for enhancing feed-forward 3D reconstruction.
\begin{itemize}
\item We propose a motion- and gravity-aware augmentation for feedforward transformer-based 3D reconstruction, improving robustness in dynamic scenes.
\item We introduce optical flow as an explicit motion cue for static--dynamic separation, and gravity direction as a global orientation prior for spatial consistency.
\item We demonstrate the effectiveness of the proposed approach through extensive experiments and ablation studies on standard multi-view reconstruction benchmarks.
\end{itemize}

\section{Related Work}
\subsection{Feed-forward 3D reconstruction}

Recent years have witnessed rapid progress in feed-forward 3D reconstruction, where models directly infer geometry from one or multiple images without test-time optimization. Early learning-based MVS systems\cite{mvsnet2018}\allowbreak\cite{yao2019recurrent}\allowbreak\cite{chen2019point}\allowbreak\cite{casmvsnet2020}\allowbreak\cite{yang2020cost} typically relied on explicit cost-volume construction and geometric reasoning, while more recent foundation-style approaches shift the emphasis toward correspondence learning with large-capacity backbones and transformers\cite{ding2022transmvsnet}\allowbreak\cite{dust3r2024}\allowbreak\cite{depthanything2024}\allowbreak\cite{wang2026pi}\allowbreak\cite{keetha2026mapanything}. In particular, transformer-based feed-forward methods such as VGGT\cite{wang2025vggt} represent scenes as collections of image patch tokens and perform cross-view reasoning through attention, achieving strong reconstruction quality and scalability. Along this line, $\pi^3$\cite{wang2026pi} further explores efficient multi-view reasoning and generalization under diverse camera configurations, while mapping-oriented systems such as MapAnything\cite{keetha2026mapanything} aim to produce geometry or map representations at scale from in-the-wild imagery, emphasizing robustness and transfer. In parallel, large-scale monocular depth pretraining (e.g., DepthAnything v3\cite{lin2025depth}) provides strong single-view geometric priors that can serve as initialization or auxiliary supervision for multi-view reconstruction and downstream 3D perception tasks. Beyond these examples, related feed-forward reconstruction and correspondence-centric paradigms include DUSt3R/MASt3R-style models\cite{dust3r2024}\allowbreak\cite{mast3r2024}\allowbreak\cite{cabon2025must3r} that unify matching and geometry prediction in a single forward pass, as well as other transformer-based MVS variants\cite{ding2022transmvsnet}\allowbreak\cite{cao2024mvsformer++} that replace hand-crafted matching pipelines\cite{schoenberger2016mvs} with learned token interactions.

Despite their success, existing feed-forward methods primarily rely on appearance-driven features and implicit geometric regularities learned from data. They generally lack explicit physical or motion priors—such as structured optical-flow cues to disentangle dynamic foregrounds or gravity direction (e.g., from IMU or geo-calibration) to serve as a global orientation reference. Consequently, these models must infer correspondences and spatial consistency purely from visual tokens, which can be brittle in dynamic scenes or under ambiguous textures \cite{zhang2025advances}. Our work complements this paradigm by injecting lightweight motion and gravity signals into feed-forward transformers, improving geometric robustness while preserving the efficiency of the inference process.

\subsection{Learning Using Privileged Information (LUPI)}

Classical supervised learning minimizes a loss function $L(h(x_i, \theta), y_i)$ over training pairs $(x_i, y_i)$, where the input space $X$ remains identical during both training and inference. In contrast, Vapnik \cite{vapnik2009new} introduced the LUPI paradigm to mimic human learning, where a "teacher" provides supplemental guidance—such as explanations or comparisons—available only during training. This transforms the training set into triplets $(x_i, x^*_i, y_i)$, where $x^*_i \in X^*$ represents \textit{privileged information} that is withheld during inference.

While early LUPI research focused on Support Vector Machines (SVMs) to predict slack variables and accelerate convergence\cite{suthaharan2016support}\allowbreak\cite{niu2012learning}, the framework has since evolved into a generalized mechanism for knowledge transfer\cite{vapnik2015learning}. LUPI has been successfully applied to diverse domains including image matting\cite{lyu2024privileged}, person re-identification\cite{yang2017person}, and healthcare analytics\cite{ganaie2022ensemble}. In this work, we integrate the LUPI paradigm with transformer-based 3D foundation models, utilizing motion and orientation priors as privileged information to guide the model toward a more robust, geometrically-aware interpretation of visual inputs.

\section{Methodology}
\subsection{Data Preprocessing}
\subsubsection{Optical flow estimation and decomposition}
We perform a preprocessing step to selectively compute the residual optical flow for the training data. For datasets composed of continuous video sequences, optical flow can generally be estimated between adjacent frames. However, when large temporal gaps or abrupt scene changes occur, flow estimation becomes unreliable. To address this issue, we first compute an image similarity score between neighboring frames and only estimate optical flow when the similarity exceeds a predefined threshold.

\begin{figure}[t]
\centering
\includegraphics[width=1.0\linewidth]{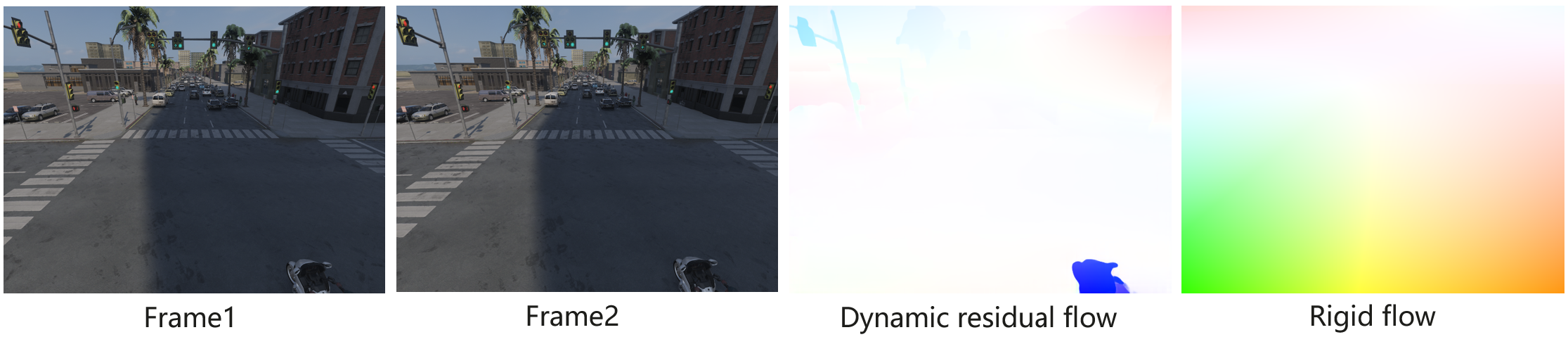}
\caption{Optical flow decomposition.}
\label{fig:opticalflow}
\end{figure}

Optical flow is computed using RAFT \cite{raft2020}, which produces dense flow fields capturing the overall motion between frames. We assume that the rigid motion induced by camera movement can be approximated by a global affine transformation, which is denoted by
\begin{equation}
  \mathbf{f}_i = A\mathbf{p}_i+\mathbf{t}
\end{equation}
where $\mathbf{f}_i$ is the optical flow of a pixel $i$. $\mathbf{p}_i$ is the $[u,v]$ coordinate of the pixel. $A$ is a $2\times2$ rotation matrix, and $\mathbf{t}$ is a $2\times1$ translation vector. $A$ and $\mathbf{t}$ can be solved by minimizing $\Sigma_i ||\mathbf{f}_i - (A \mathbf{p}_i + \mathbf{t})||^2$. Pixels consistent with the affine transformation are treated as static background, while deviations from this model indicate dynamic regions. The residual flow is defined as the difference between the original optical flow and the estimated affine transformation, thereby isolating motion components that cannot be explained by the global camera motion model. Figure~\ref{fig:opticalflow} shows two adjacent input frames together with the decomposed rigid flow and the resulting residual.

Importantly, residual flow does not solely correspond to independently moving objects. The affine transformation provides only a global approximation of camera-induced motion and therefore cannot perfectly model perspective projection effects in scenes with significant depth variation. In regions with large depth gradients, structural discontinuities, or partial occlusions, the global motion model becomes insufficient even under predominantly static conditions. Consequently, these regions yield non-negligible residual flow. Such areas typically coincide with geometrically challenging regions for multi-view aggregation, as illustrated by the traffic light region in Fig.~\ref{fig:opticalflow}. Therefore, residual flow can also be interpreted as a signal highlighting deviations from the global motion model, which often require more careful geometric reasoning during reconstruction.

\subsubsection{Gravity direction estimation}
We estimate 3D gravity vectors in the camera coordinate system for all training images using the pretrained GeoCalib model \cite{veicht2024geocalib}. The model is applied offline to each image to extract its corresponding gravity direction.

GeoCalib first predicts a dense perspective field from a single image, which encodes geometric cues such as vanishing directions and horizon structure. Based on this representation, gravity direction is recovered through a Levenberg–Marquardt optimization procedure. To ensure reliability, we retain only those gravity estimates for which the optimization converges within a predefined number of iterations. In practice, non-convergence occurs only in a very small fraction of images. Samples for which the optimization fails to converge are excluded from gravity supervision. The resulting gravity vectors, expressed in the camera coordinate system, are used as orientation priors during training.

\subsection{Privileged Information Injection}

To incorporate additional geometric cues during training, we introduce a privileged information injection mechanism that integrates residual optical flow and gravity direction into the transformer-based reconstruction backbone. These signals are encoded into token-aligned representations and injected into the aggregation process with minimal modification to the original architecture.

\paragraph{Residual Flow Encoding.}
Residual optical flow is first partitioned into patches using the same patchification scheme as the input images. For each patch, we compute a compact statistical representation consisting of five parameters: the mean flow components $\mu_x, \mu_y$, the log-variance terms $\log(\sigma_{xx}+\epsilon), \log(\sigma_{yy}+\epsilon)$, and the normalized covariance $\sigma_{xy}/\sqrt{\sigma_{xx}\sigma_{yy}+\epsilon}$.

These statistics summarize the motion distribution within each patch and provide a lightweight representation of local motion dynamics.

The resulting 5-dimensional vector is passed through a three-layer MLP, referred to as the \textit{patch encoder}, to produce a residual-flow embedding with the same dimensionality as the visual patch tokens extracted by the DINO backbone. The encoded residual-flow feature is then scaled by a learnable weight $W_{\text{flow}}$ and added to the corresponding patch token:

\begin{equation}
\tilde{t}_{patch} = t_{patch} + W_{\text{flow}} \cdot f_{flow}.
\end{equation}

This additive fusion allows motion information to be integrated with minimal perturbation to the existing embedding space while enabling the network to leverage dynamic cues during multi-view aggregation.

\paragraph{Gravity Encoding.}
In addition to motion cues, we incorporate scene orientation information through gravity direction. The gravity vector is first normalized and then encoded using a two-layer MLP, referred to as the \textit{gravity encoder}, producing a representation aligned with the dimensionality of camera tokens. The encoded gravity feature is scaled by a learnable parameter $W_{\text{grav}}$ and fused with the camera token:

\begin{equation}
\tilde{t}_{cam} = t_{cam} + W_{\text{grav}} \cdot f_{grav}.
\end{equation}

The specific implementation of this fusion varies depending on the model architecture. For $\pi^3$, since there are no specialized camera tokens, we define the camera tokens as the aggregator encoder's output embeddings that feed into the camera head. The gravity feature corresponding to a specific image is fused with all constituent patch tokens for the image.

For VGGT, we modify the gravity input to align with its reference-frame design. The first camera token in the sequence receives no gravity information, as it serves as a fixed reference for camera parameters. All subsequent tokens are fused with a concatenated gravity vector containing both their own view's orientation and that of the first view. This ensures the gravity features remain aware of the reference frame with minimal learning overhead.

This design allows the model to incorporate global orientation priors that help stabilize camera estimation and improve geometric consistency across views.

\begin{figure}[t]
\centering
\includegraphics[width=0.8\textwidth]{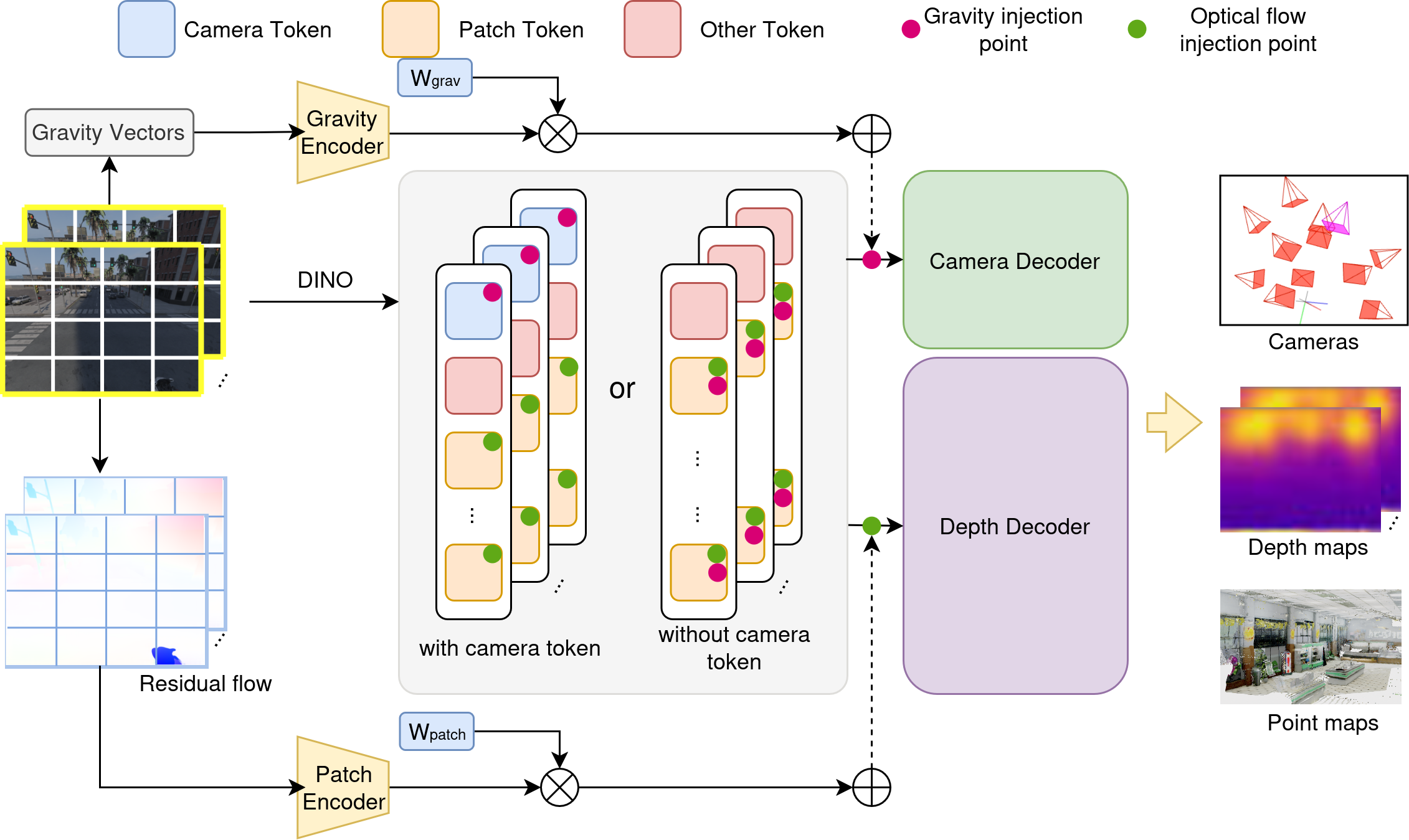}
\caption{Architecture Overview.}
\label{fig:pipeline}
\end{figure}

\paragraph{Injection Locations.}
The encoded privileged information can be injected at different stages of the transformer-based aggregation module, as illustrated in Fig.~\ref{fig:pipeline}. Specifically, residual flow features are fused with patch tokens, while gravity embeddings are combined with camera tokens when such tokens are present.

For reconstruction architectures that do not explicitly maintain camera tokens (e.g., $\pi^3$), gravity embeddings are instead injected before the camera decoder to influence the camera prediction stage.

The framework supports multiple potential injection points throughout the aggregator layers, as shown in Fig.~\ref{fig:pipeline}. Empirically, our experiments show that the optimal injection position depends on the backbone architecture, with different models favoring different stages of the aggregator.

\subsection{Finetuning}

Our training objectives include local and global point maps, log-space depth maps, pixel ray directions, and pairwise camera translations and rotations. Additionally, for synthetic datasets, we incorporate surface normal and depth-gradient signals to provide finer geometric supervision. All loss terms are computed using the Generalized Robust Loss \cite{barron2019general}, with the final objective being a weighted sum of these individual components. We provide detailed formulae and hyperparameter configurations in the supplementary materials. 

Following the principles of Learning Using Privileged Information (LUPI), we employ a two-stage finetuning strategy designed to extract reliable privileged features and subsequently train the model to interpret and hallucinate them during inference.

\paragraph{Stage 1: Privileged Feature Alignment.}
The primary objective of the first stage is to develop a robust privileged information encoder and adapt the model's decoders to utilize it. We freeze the model layers preceding the information injection point, focusing training exclusively on the privileged encoder (i.e. residual flow or gravity encoder) and the downstream decoder layers. This isolation ensures the new modality encoder receives sufficient gradient signal to provide meaningful features while the decoders learn to leverage this injected information. By freezing the early encoder layers, the privileged features are forced to align with the existing distribution of patch and camera tokens, facilitating better fusion. To minimize architectural disturbance, we also freeze unrelated decoders and heads; for instance, the VGGT camera decoder remains frozen in stage 1 when finetuning with residual flow as the privileged modality.

\paragraph{Stage 2: Knowledge Distillation and Hallucination.}
As privileged modalities are unavailable at inference time under the LUPI paradigm, the second stage shifts toward guiding the primary model to converge faster and learn more robustly. With the privileged encoder and decoders now prepared, we unfreeze the entire model and introduce aggressive modality dropout. This mechanism creates a shortcut for the finetuning process, forcing the unfrozen model encoder to "hallucinate" the missing privileged features when they are dropped out. In this framework, the privileged information encoder acts as a teacher during standard iterations, while the model encoder serves as the student during dropout iterations. The student learns to extract high-level features related to motion or gravity cues directly from visual inputs, effectively filling the gap left by the privileged modality to maintain the performance gains established in stage 1.

\section{Experiments}

\subsection{Datasets}
We apply dataset-level train-test split using 13 high-quality datsets. Among them, 11 are used for training and 2 are used for benchmarking. For our LUPI method, gravity vectors are pre-computed for all datasets, while dynamic flow maps are computed for 6 of them where continuous scene frames are available.
A wide range of indoor and outdoor scenes are covered in the datasets either synthetic or captured in the wild. In each training epoch, we randomly sampled a fixed number of scenes from each training dataset: 1,000 in Spring, 2,750 in UnrealStereo4K, and 26,250 in the other 9 datasets. We preprocess the datasets into WAI format\cite{keetha2026mapanything} for the ease of training and benchmarking.

\begin{table}[t]
\centering
\tiny
\caption{Datasets used for training and testing.}
\label{tab:datasets}
\begin{tabular}{l c c c}
\toprule
\textbf{Dataset} & \textbf{\#images} & \textbf{\#images w OF} & \textbf{\%images w OF} \\
\midrule
\multicolumn{4}{l}{\textbf{Datasets for training}} \\
BlendedMVS\cite{yao2020blendedmvs} & 230,284 & 0 & 0 \\
Mapillary Planet-Scale Depth\cite{antequera2020mapillary} & 1,688,784 & 0 & 0 \\
Spring\cite{Mehl2023_Spring} & 34,000 & 11,853 & 34.9 \\
UnrealStereo4K\cite{Tosi2021CVPR} & 16,400 & 0 & 0 \\
Aria Synthetic Environments\cite{avetisyan2024scenescript} & 11,765,050 & 10,647,370 & 90.5 \\
DL3DV-10K\cite{ling2024dl3dv} & 5,178,105 & 0 & 0 \\
Dynamic Replica\cite{karaev2023dynamicstereo} & 337,800 & 244,027 & 72.24 \\
MegaDepth\cite{MegaDepthLi18} & 41,311 & 0 & 0 \\
MVS-Synth\cite{DeepMVS} & 12,000 & 10,836 & 90.3 \\
ParallelDomain-4D\cite{vanhoorick2024gcd} & 1,448,106 & 1,159,273 & 80.1 \\
SAIL-VOS 3D\cite{HuCVPR2021} & 79,953 & 76,386 & 95.5 \\
\midrule
\multicolumn{4}{l}{\textbf{Datasets for testing}} \\
ETH3D\cite{schoeps2017cvpr} & 647 & / & / \\
TartanAirV2-WB\cite{zhang2025ufm} & 727,360 & / & / \\
\bottomrule
\end{tabular}
\end{table}

\subsection{Implementation Details} 

We implement our method using the official pretrained checkpoints of $\pi^3$ \cite{wang2026pi} and VGGT \cite{wang2025vggt}, fine-tuning them with the proposed motion and orientation priors. The injection points for privileged information are tailored to each architecture: for VGGT, information is injected into the 12th layer of the aggregator. In contrast, as $\pi^3$ lacks explicit camera tokens, we inject additional modalities after the aggregator but before the specific transformer decoders to maintain architectural isolation. Specifically, gravity features are injected before the camera decoder, while residual flow features are integrated before the point and confidence decoders. A detailed analysis of these injection points is provided in Section \ref{ablation}.

All models are fine-tuned for a total of 10 epochs. For single-modality configurations, Stage 1 (Privileged Alignment) spans the first 2 epochs, while the remaining 8 epochs are dedicated to Stage 2 (Knowledge Distillation). During Stage 2, we employ a scene-wise modality dropout rate fixed at 0.5 to encourage the model encoder to hallucinate privileged features. To ensure stable gradient flow to the privileged encoders, the scaling weights $W_{\text{flow}}$ and $W_{\text{grav}}$ are initialized and fixed at 1.0 during the first two epochs.

For experiments involving both modalities, we utilize a sequential multi-stage fine-tuning schedule. The model first undergoes Stages 1 and 2 for optical flow injection during the initial 5 epochs. In the subsequent 5 epochs (Stages 3 and 4), we introduce gravity injection using the same two-stage protocol while maintaining the presence of the flow injection. This sequential curriculum allows the model to stabilize one privileged modality before integrating the second, ensuring coherent multi-modal feature fusion.

\subsection{Dense Multi-View Evaluation}

\begin{figure}[p]
    \centering
    \includegraphics[width=1\textwidth]{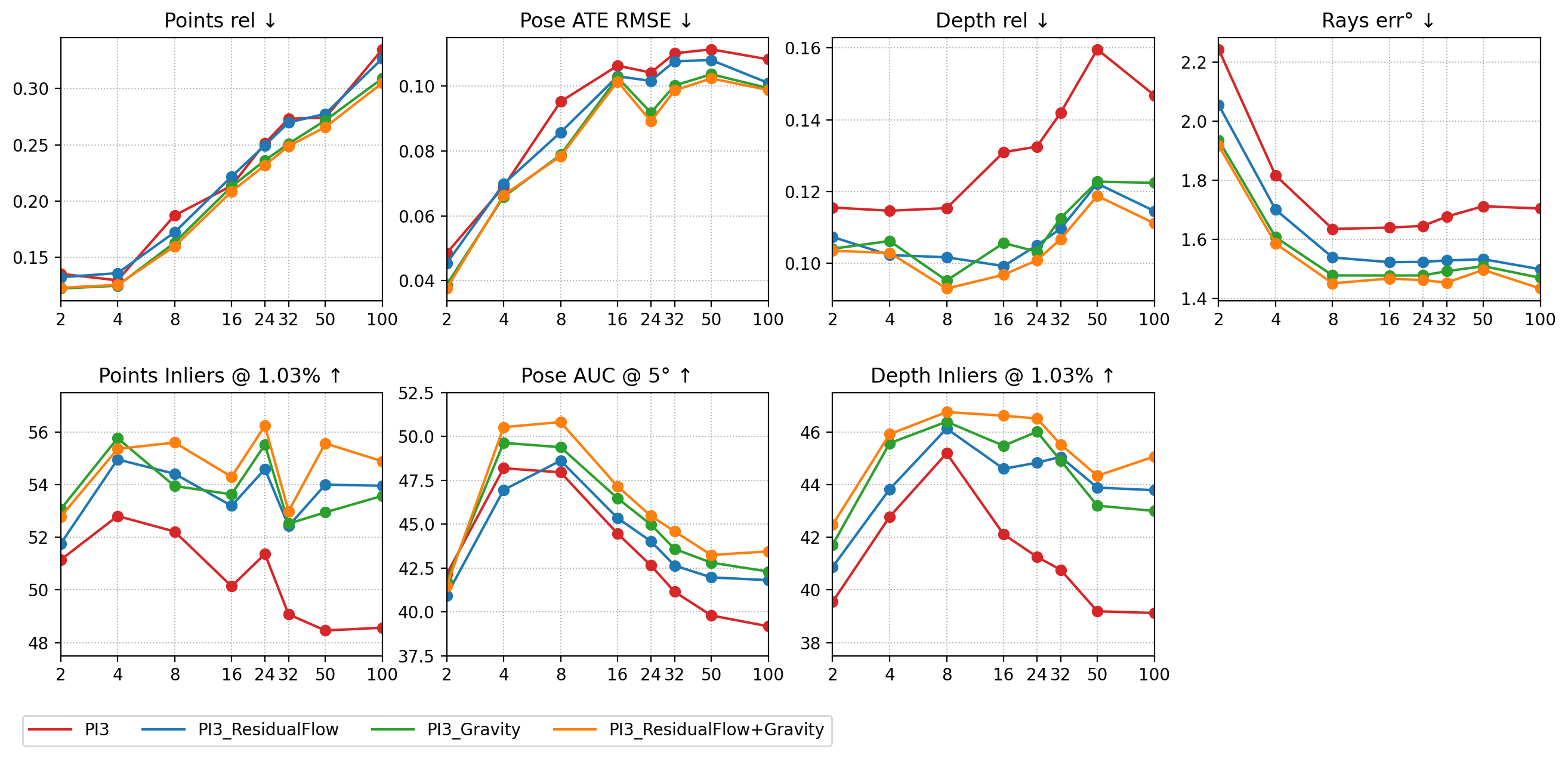}
    \caption{$\pi^3$ injection results}
    \label{fig:pi3_injection}
\end{figure}

\begin{figure}[p]
    \centering
    \includegraphics[width=1\textwidth]{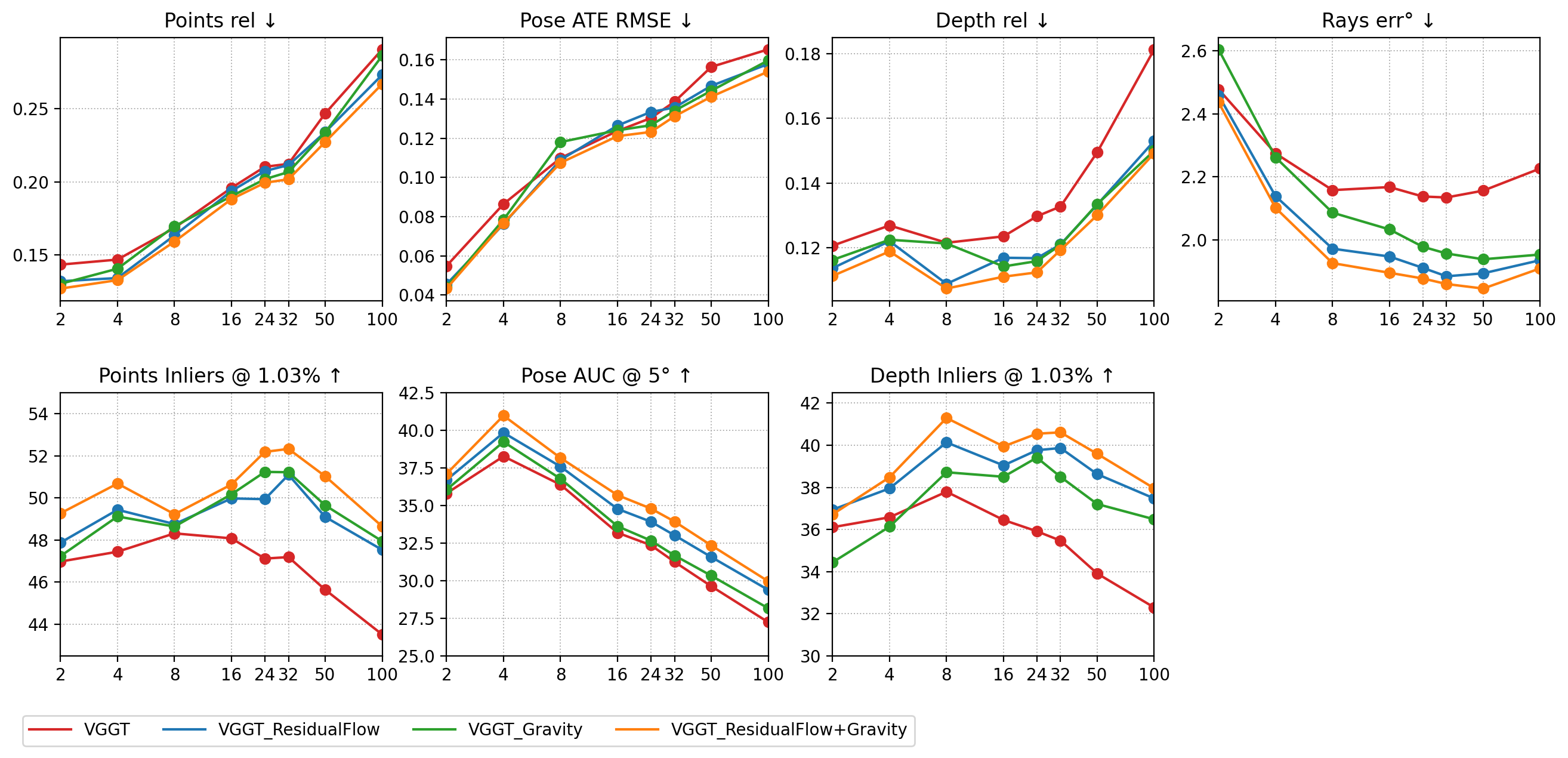}
    \caption{VGGT injection results}
    \label{fig:vggt_injection}
\end{figure}

We evaluate our method under the dense multi-view protocol using both $\pi^3$ and VGGT backbones. Figures \ref{fig:pi3_injection} and \ref{fig:vggt_injection} report performance across a varying number of input views.

On both architectures, injecting \textit{residual flow} consistently outperforms the baselines across all key metrics, including depth relative error, ray alignment, and inlier-based statistics. These gains confirm that explicitly modeling dynamic motion is highly beneficial for feed-forward multi-view reconstruction. Similarly, the introduction of \textit{gravity priors} yields results comparable to residual flow, further validating the role of global orientation cues in enhancing structural alignment. Crucially, models fine-tuned with both modalities outperform those using any single injection, demonstrating that these priors are complementary rather than redundant or contradictory.

Notably, the performance gap between our method and the baseline widens as the number of input views increases. In dense settings, cross-view aggregation becomes the dominant factor, which simultaneously amplifies geometric inconsistencies caused by independently moving objects or misaligned camera orientations. By effectively suppressing these inconsistencies, motion and gravity awareness leads to increasingly pronounced gains as view redundancy grows. This trend indicates that our approach actively enhances cross-view geometric aggregation rather than merely compensating for sparse observations. 

Furthermore, the consistency of these improvements across both $\pi^3$ and VGGT demonstrates that our proposed learning framework generalizes effectively across different feed-forward reconstruction backbones without requiring specialized architectural modifications.

\clearpage

\subsection{Ablation Studies}\label{ablation}

\subsubsection{Residual vs. Full Flow Injection}
To examine the necessity of motion decomposition, we conduct experiments by injecting both full optical flow and residual flow into VGGT and $\pi^3$. Table \ref{tab:res_full_injection} reports the reconstruction performance of each variant under the 8-view setting.

Empirically, injecting full optical flow yields performance comparable to the baseline, with slight degradation observed in some metrics. In contrast, injecting residual flow consistently achieves the best performance across both backbones.

A possible explanation is that full optical flow contains both camera-induced rigid motion and dynamic motion. But with the random set of views sampling for 3D vision models' input, there is no guarantee of the sequence of the view nor the existance of the next view whose camera relate to the source view in the way expressed by rigid motion. In contrast, By explicitly removing the rigid component, residual flow isolates motion signals that deviate from the global motion due to the camera. These signals often correspond to independently moving objects or geometrically challenging regions, making them more readily helpful for 3D reconstruction.

\begin{table}[t]
\centering
\tiny
\caption{Residual vs. Full Flow Injection.}
\label{tab:res_full_injection}
\begin{tabular}{l l c c c c}
\toprule
\textbf{Backbone} & \textbf{Flow Type} & 
\textbf{Depth rel} $\downarrow$ & 
\textbf{Rays err$^\circ$} $\downarrow$ & 
\textbf{Pose AUC} $\uparrow$ & 
\textbf{Depth Inliers} $\uparrow$ \\
\midrule

\multirow{2}{*}{$\pi^3$} 
& Residual flow & \textbf{0.107} & \textbf{2.055} & 40.94 & \textbf{40.87} \\
& Full flow     & 0.113 & 2.235 & \textbf{42.19} & 38.76 \\
& No injection  & 0.116 & 2.242 & 42.14 & 39.55 \\
\midrule

\multirow{2}{*}{VGGT} 
& Residual flow & \textbf{0.114} & \textbf{2.459} & \textbf{36.69} & \textbf{36.94} \\
& Full flow     & 0.124 & 2.491 & 35.46 & 35.44 \\
& No injection  & 0.121 & 2.478 & 35.80 & 36.11 \\
\bottomrule
\end{tabular}
\end{table}

\subsubsection{Layer-wise Injection Analysis}
To determine the optimal depth for incorporating motion and orientation priors, we evaluate feature injection at three distinct stages of the reconstruction backbone: \textit{Early} (input image patch features), \textit{Middle} (intermediate aggregator layers), and \textit{Late} (aggregator output). Performance benchmarks for 8-view scenes are summarized in Table \ref{tab:layer_injection}. Our results reveal a consistent architectural preference: middle-layer injection performs best on most metrics for VGGT, while late-layer injection is optimal for $\pi^3$.

\paragraph{VGGT (Middle-Layer).} At this depth, patch and camera tokens have developed sufficient global geometric awareness to fuse effectively with privileged features. Unlike late injection, this placement allows subsequent transformer blocks to refine the fused representations through cross-view attention, ensuring deeper integration before reaching the decoders. Furthermore, VGGT's explicit camera tokens provide natural modality isolation: gravity features fused with camera tokens are directed exclusively to the camera head, while residual flow features in patch tokens are reserved for depth-related decoders.

\paragraph{$\pi^3$ (Late-Layer).} The absence of explicit camera tokens in $\pi^3$ means the aggregator output is shared across all heads. Middle-layer injection causes privileged information to leak into unrelated decoders (e.g., gravity features influencing the point decoder). Injecting at the aggregator output instead allows for precise, head-specific integration, providing the isolation necessary for superior performance.

\paragraph{Early Injection.} This configuration consistently yields the poorest results across all cases. Raw input features lack the cross-view geometric context required to interpret motion priors. Additionally, this placement forces the pretrained image encoder—which lacks multi-view awareness—to handle the complex task of feature hallucination, a burden that undermines the LUPI learning shortcut and prevents the model from effectively leveraging the privileged information.

\begin{table}[t]
\centering
\tiny
\caption{Layer-wise injection analysis on $\pi^3$ and VGGT.}
\label{tab:layer_injection}
\setlength{\tabcolsep}{3pt}
\begin{tabular}{l l c c c c}
\toprule
\textbf{Backbone} & \textbf{Injection Layer} & 
\textbf{Depth rel} $\downarrow$ & 
\textbf{Rays err$^\circ$} $\downarrow$ & 
\textbf{Pose AUC} $\uparrow$ & 
\textbf{Depth Inliers} $\uparrow$ \\
\midrule

\multirow{3}{*}{$\pi^3(ResFlow)$} 
& Early (Input)            & 0.115 & 1.714 & 46.19 & 44.43  \\
& Middle (Layer 17)        & 0.110 & 1.621 & 47.87 & 45.85 \\
& \textbf{Late (Output)}   & \textbf{0.102} & \textbf{1.539} & \textbf{48.62} & \textbf{46.13}  \\
\midrule

\multirow{3}{*}{$\pi^3(Grav)$} 
& Early (Input)            & 0.104 & 1.707 & 43.03 & 44.08  \\
& Middle (Layer 17)        & 0.096 & 1.561 & 44.38 & 45.67 \\
& \textbf{Late (Output)}   & \textbf{0.095} & \textbf{1.478} & \textbf{49.39} & \textbf{46.39}  \\
\midrule

\multirow{3}{*}{VGGT(ResFlow)} 
& Early (Input)                & 0.129 & 2.369 & 34.16 & 36.56  \\
& \textbf{Middle (Layer 11)}   & \textbf{0.109} & \textbf{1.973} & \textbf{37.58} & \textbf{40.13} \\
& Late (Output)   & 0.114 & 2.167 & 37.52 & 38.07 \\
\midrule

\multirow{3}{*}{VGGT(Grav)} 
& Early (Input) & 0.123 & 2.362 & 36.36 & 38.18  \\
& \textbf{Middle (Layer 11)} & \textbf{0.121} & \textbf{2.087} & \textbf{36.77} & 38.72 \\
& Late (Output)   & 0.118 & 2.188 & 37.44 & \textbf{38.85} \\
\bottomrule
\end{tabular}
\end{table}

\subsubsection{Gravity Prior Analysis}

\begin{figure}[t]
\centering
\includegraphics[height=0.25\linewidth]{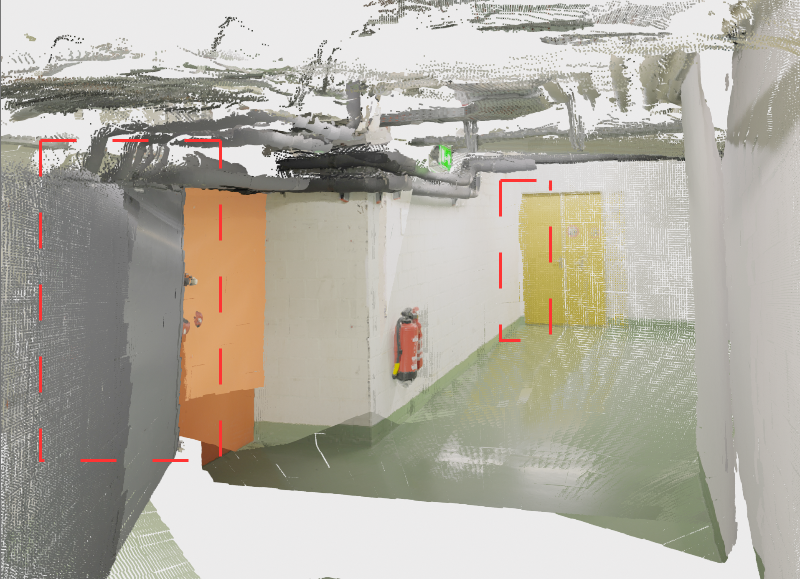}
\includegraphics[height=0.25\linewidth]{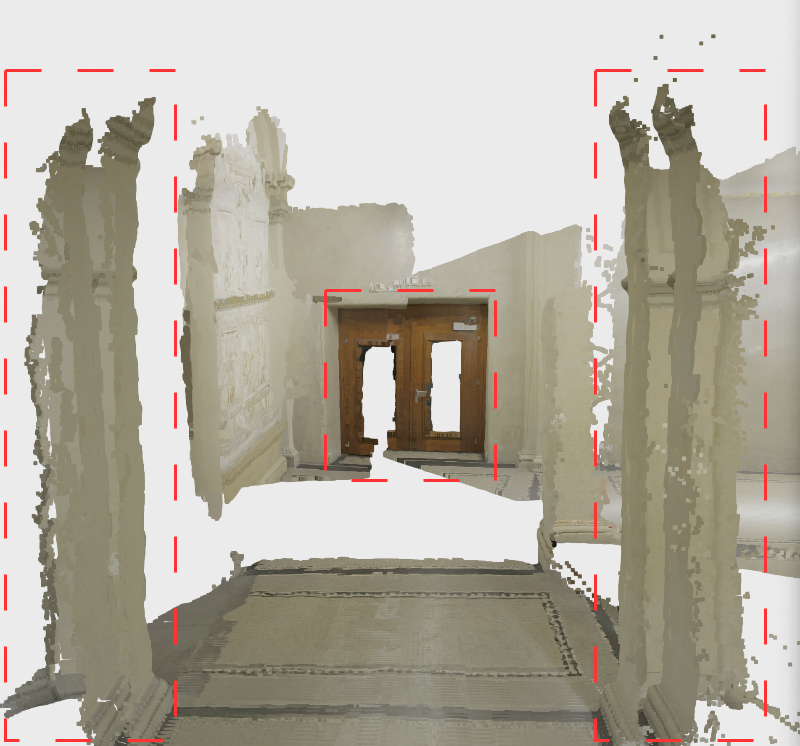}
\includegraphics[height=0.25\linewidth]{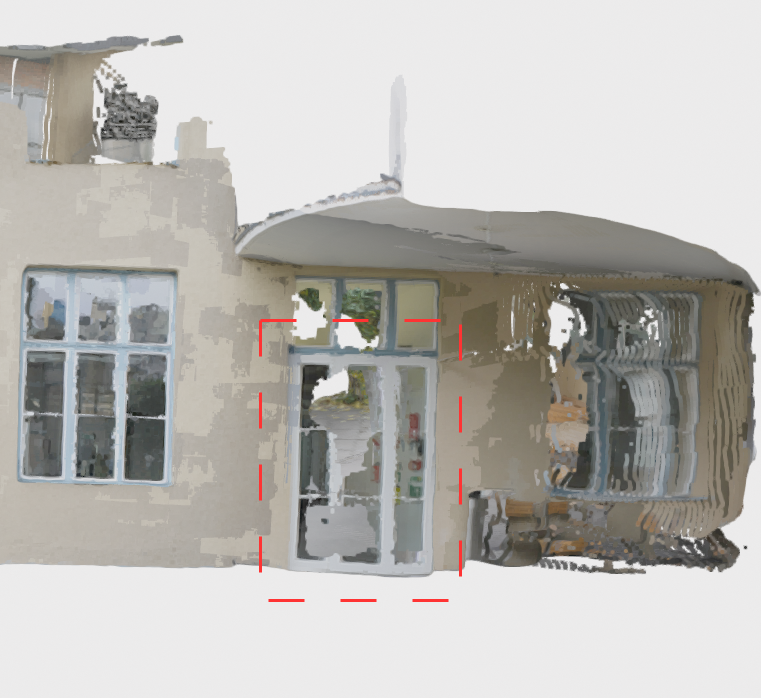}

\includegraphics[height=0.25\linewidth]{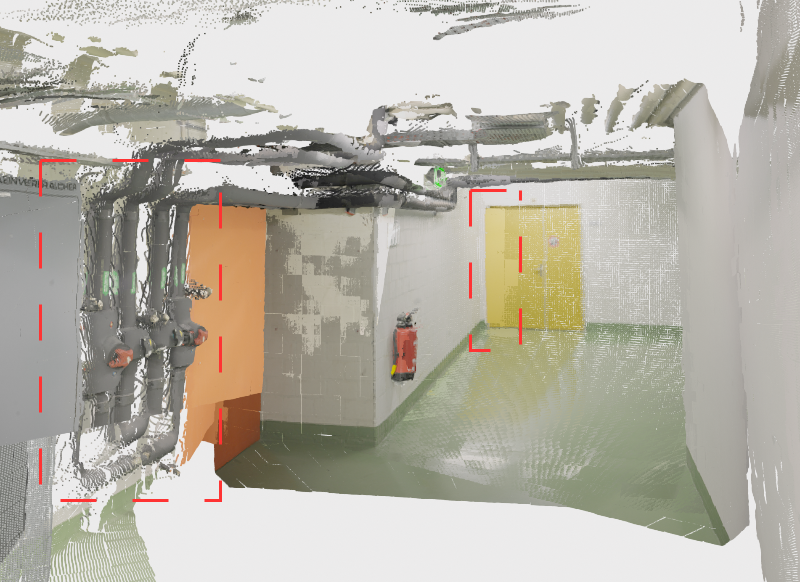}
\includegraphics[height=0.25\linewidth]{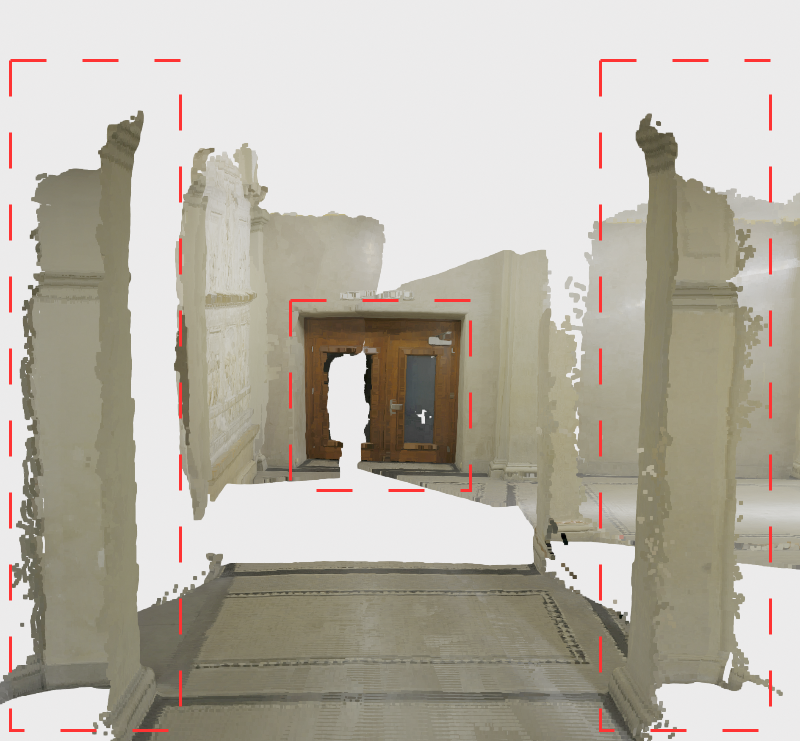}
\includegraphics[height=0.25\linewidth]{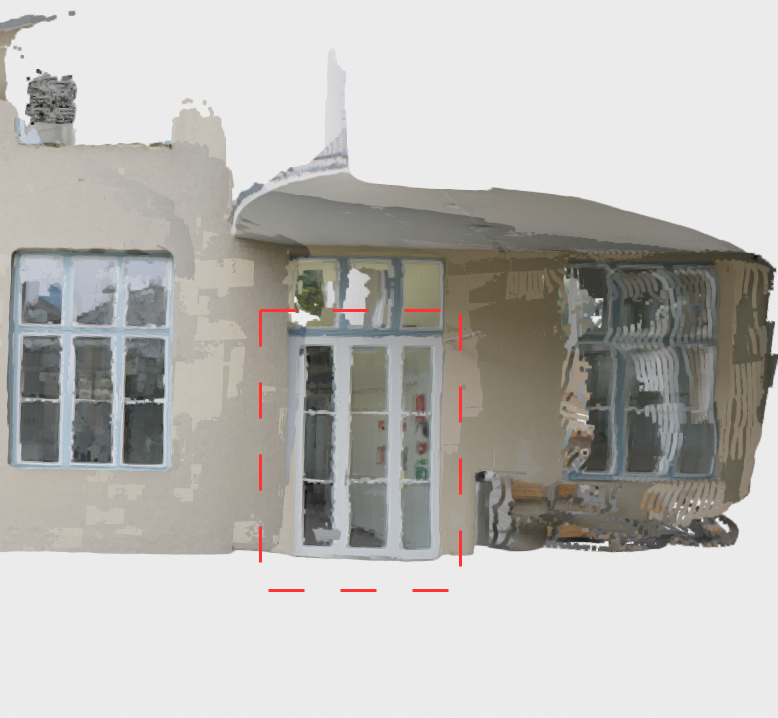}
\caption{Qualitative comparison with and without gravity injection. In each pair, the upper image shows reconstruction without gravity injection, and the lower image shows reconstruction with gravity injection.}
\label{fig:pointmap}
\end{figure}

To evaluate the effect of gravity injection, we train two models under identical settings: one with gravity prior and one without it. Figure~\ref{fig:pointmap} shows the reconstructed point maps predicted by the two models. The model trained with gravity prior produces more stable global structures, where ground surfaces appear flatter and walls are more consistently aligned with the vertical direction.

This observation suggests that gravity prior mainly improves reconstruction by enforcing a global orientation reference rather than altering local geometric predictions. By anchoring the reconstruction to a physically meaningful direction, gravity helps reduce structural drift and leads to more geometrically coherent scene reconstruction.

\subsection{Analysis}

We discuss the proposed priors from three perspectives: view-density behavior, architecture-dependent injection preference, and prior complementarity.

First, the gain over the baseline increases with the number of input views. This trend indicates that the guidance of motion and gravity priors does not simply compensate for sparse observations; instead, they improve the consistency of cross-view geometric aggregation. As view density grows, dynamic regions introduce stronger multi-view conflicts, and residual flow helps suppress these inconsistencies.

Second, the optimal injection stage is backbone-dependent. For $\pi^3$, late-stage injection performs best, suggesting that decoder-side integration better preserves modality isolation when no explicit camera token is available. For VGGT, middle-layer injection is generally more effective, likely because privileged cues can be fused earlier with geometry-aware tokens and further refined by subsequent cross-view transformer blocks. In both backbones, early-stage injection is consistently weaker, implying that shallow appearance-dominant features are less suitable for privileged geometric guidance.

Third, residual flow and gravity priors contribute at complementary structural scales. Residual flow primarily improves local dynamic consistency by emphasizing non-rigid motion components, while gravity provides a global orientation anchor that stabilizes scene-level structure. Their combination improves geometric coherence without changing the feed-forward inference paradigm.

\section{Limitations}
A limitation of our current formulation lies in the use of rigid optical flow components. In principle, rigid flow primarily reflects the apparent motion induced by camera ego-motion, implicitly encoding information about the underlying translational velocity and angular velocity of the camera. Such signals can be highly informative for stabilizing geometry and improving motion-aware reasoning.

However, most existing feedforward multi-view reconstruction frameworks, including VGGT, are designed for static-scene reconstruction and operate on unordered sets of input images, without explicit temporal ordering or timestamp information. Under this setting, the rigid flow component does not provide a consistent notion of camera dynamics across views, and consequently offers limited benefit for reconstruction performance in our experiments.

We believe that the velocity and angular motion cues embedded in rigid flow remain a valuable source of physical information. An important direction for future work is to extend feedforward reconstruction toward streaming or temporally coherent settings, where sequential inputs and ego-motion estimates can be explicitly modeled. Incorporating rigid motion cues in such a framework may further improve robustness and enable more principled dynamic scene reconstruction.

\section{Conclusion}
In this work, we presented a motion- and gravity-aware framework for feed-forward multi-view 3D reconstruction under the Learning Using Privileged Information (LUPI) paradigm. During training, residual flow and gravity cues are used as privileged supervision signals; during inference, the model operates with standard visual inputs only, while retaining the geometric benefits transferred from privileged modalities.

Extensive experiments show consistent improvements in reconstruction accuracy and robustness, especially in dynamic scenes, under ambiguous geometry, and across challenging camera motion. The results further indicate that the optimal injection stage is architecture-dependent, highlighting the importance of aligning privileged-feature integration with backbone design.

Overall, this study demonstrates that LUPI is an effective and practical mechanism for introducing motion and physical priors into feed-forward 3D reconstruction without sacrificing inference efficiency. We hope this perspective motivates future work on temporally coherent and physically grounded feed-forward reconstruction with richer privileged training signals.

\clearpage
\nocite{wang2025moge,Yu2022MonoSDF,Loshchilov2017DecoupledWD}
\bibliographystyle{unsrt}
\bibliography{egbib}
\clearpage
\appendix

\section{Finetuning Details}

We employ the WAI\_format \cite{keetha2026mapanything} for dataset pre-processing and fine-tuning. Specifically, we determine the co-visibility between all image pairs within a scene by evaluating the re-projection error derived from ground-truth depth maps and camera poses. During fine-tuning, input sequences are curated by sampling images that maintain a co-visibility of at least 25\% relative to at least one other view in the sequence; this ensures consistent spatial overlap across the chosen views. For each input sequence, we compute local and global point maps, ray directions and distances, camera translations, and rotations (represented as quaternions), alongside a per-view confidence map.

All loss terms are calculated using the robust regression loss \cite{barron2019general} with a shape parameter of $0.5$ and a scaling parameter of $0.25$ to mitigate the influence of outliers. We apply a logarithmic transformation to the ground truth and model outputs for point maps and $z$-depth values prior to computing the regression loss on their differences; notably, no loss terms are calculated for ray distances or confidence maps. When high-quality, smooth ground-truth point maps are available, we also integrate normal loss \cite{wang2025moge} and gradient matching loss \cite{Yu2022MonoSDF} into our objective function. To account for noise and extreme values inherent in ``in-the-wild'' real-world datasets, we apply confidence weights to the global point map losses and exclude the top 5\% of errors as outliers for local point maps and VGGT depth losses. To align with the reference-view-free architecture of $\pi^3$, we compute pairwise relative pose losses rather than direct differences between predicted and ground-truth global poses. The same 5\% outlier exclusion logic is applied to all three losses for $\pi^3$.

The loss terms for the different configurations and dataset types are expressed as follows:

% Case 1: VGGT - Synthetic Datasets
\begin{equation}
    \mathcal{L}_{\text{VGGT}}^{\text{syn}} = L_{pt} + L_{z} + L_{ray} + L_{pose} + 3L_{norm} + 3L_{grad}
\end{equation}
% Case 2: VGGT - Real-World Datasets
\begin{equation}
    \mathcal{L}_{\text{VGGT}}^{\text{real}} = W_{conf} \cdot L_{pt, glob} + M_{top5\%} \cdot (L_{pt, loc} + L_{z}) + L_{ray} + L_{pose}
\end{equation}
% Case 3: \pi^3 - Synthetic Datasets
\begin{equation}
    \mathcal{L}_{\pi^3}^{\text{syn}} = L_{pt} + L_{z} + L_{ray} + L_{rel} + 3L_{norm} + 3L_{grad}
\end{equation}
% Case 4: \pi^3 - Real-World Datasets
\begin{equation}
    \mathcal{L}_{\pi^3}^{\text{real}} = M_{top5\%} \cdot (L_{pt, glob} + L_{pt, loc} + L_{z}) + L_{ray} + L_{rel}
\end{equation}

where:
\begin{itemize}
    \item $L_{pt}$, $L_{z}$, and $L_{ray}$ represent the robust regression losses for point maps, z-depth values, and ray directions, respectively.
    \item $L_{pose}$ denotes the global camera pose loss (translation and rotation) used for VGGT.
    \item $L_{rel}$ denotes the pairwise relative pose loss used for the reference-view-free $\pi^3$ architecture.
    \item $L_{norm}$ and $L_{grad}$ are the normal and gradient matching losses, applied only when smooth synthetic ground truth is available.
    \item $W_{conf}$ is the per-pixel confidence weight applied to global point map predictions.
    \item $M_{top5\%}$ is the binary outlier mask that excludes the highest 5\% of errors to ensure robustness against noise in real-world data.
\end{itemize}

Optimization is performed using the AdamW optimizer \cite{Loshchilov2017DecoupledWD} with hyperparameters set to $\beta_{1}=0.9$, $\beta_{2}=0.95$, and a weight decay of $0.05$. We employ a differentiated learning rate strategy: the peak learning rate is set to $5 \times 10^{-6}$ for the DINOv2 image encoder, $5 \times 10^{-5}$ for the gravity and flow encoders when present, and $1 \times 10^{-5}$ for all other architectural components. The training follows a structured schedule throughout Stage 1, consisting of a 20\% linear warmup to reach the peak learning rate, followed by a half-cycle cosine decay to a final value two orders of magnitude lower.

\section{Additional Ablation Results}
\begin{figure}[t]
\centering
\includegraphics[width=1\textwidth]{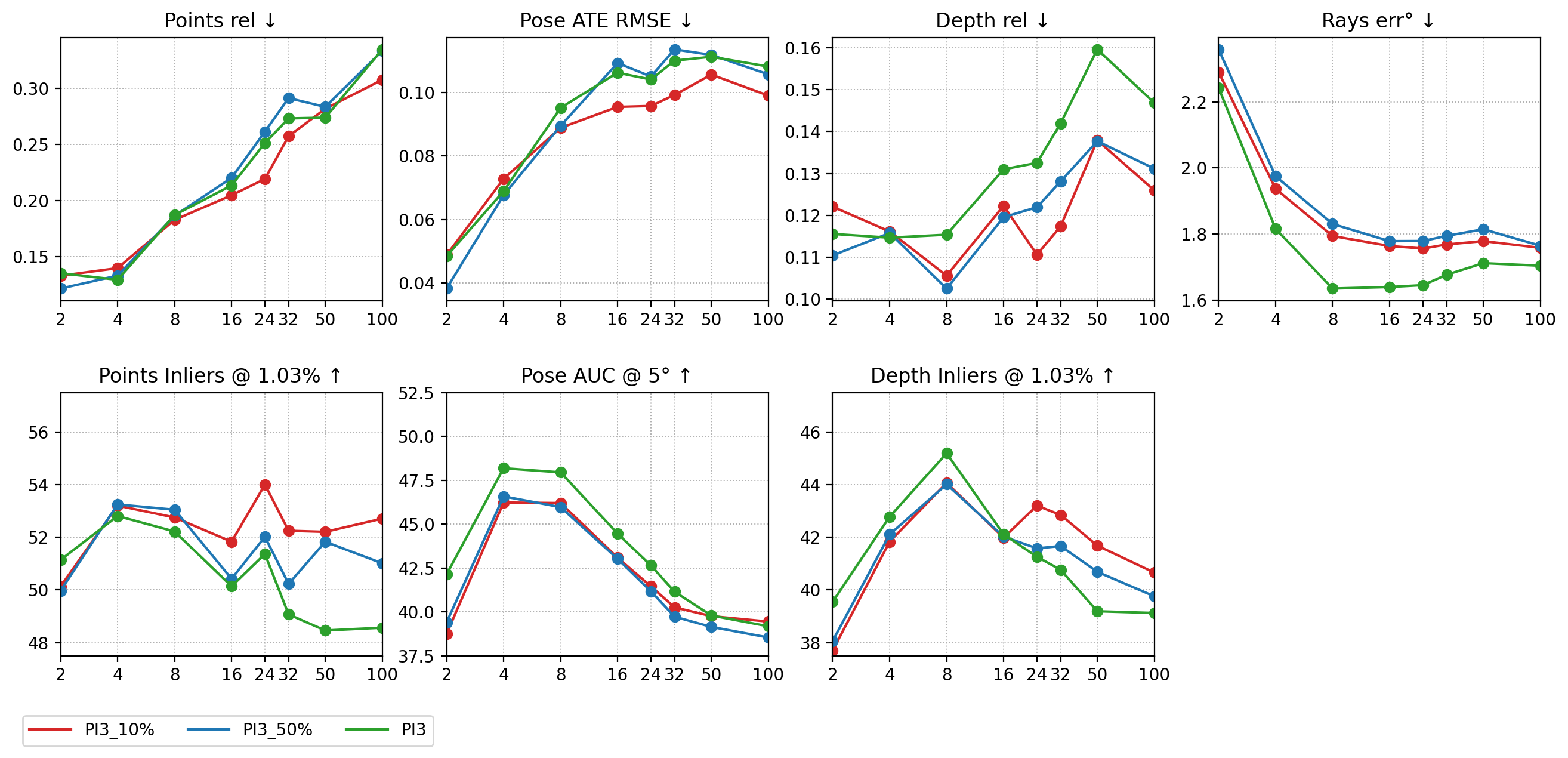}
\caption{Performance of the $\pi^3$ model fine-tuned with varying effective dataset scales. PI3\_10\% denotes training with 10\% of scenes sampled per epoch over a $10\times$ total epoch duration, while PI3\_50\% utilizes 50\% of scenes over a $2\times$ duration to maintain constant total training steps.}
\label{fig:pi3_ablation}
\end{figure}

\begin{figure}[t]
\centering
\includegraphics[width=1\textwidth]{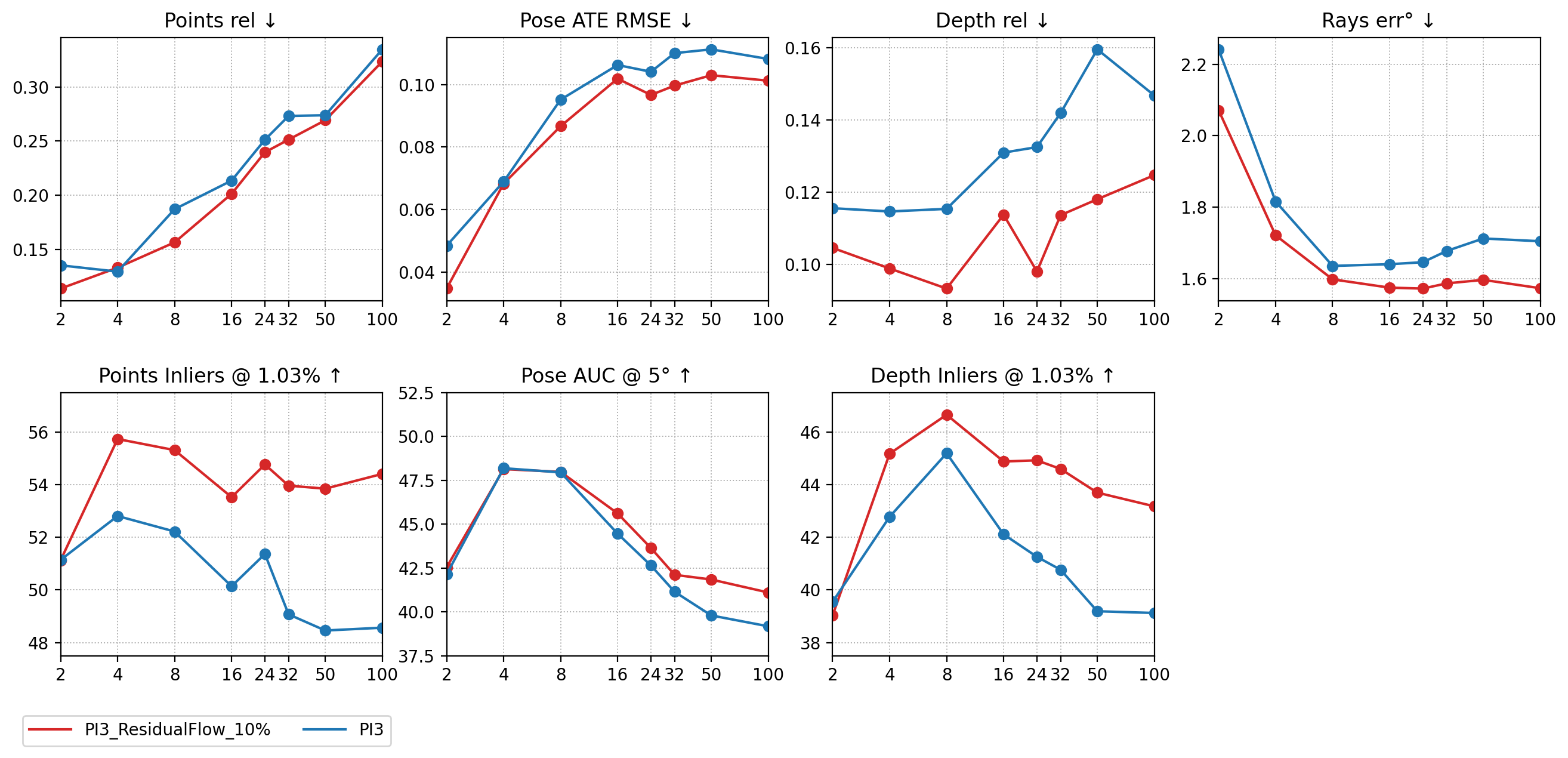}
\caption{Comparison of $\pi^3$ performance under default fine-tuning settings versus fine-tuning with a 10\% effective dataset size augmented by residual flow injection.}
\label{fig:pi3_comparison}
\end{figure}

\begin{figure}[t]
\centering
\includegraphics[width=1\textwidth]{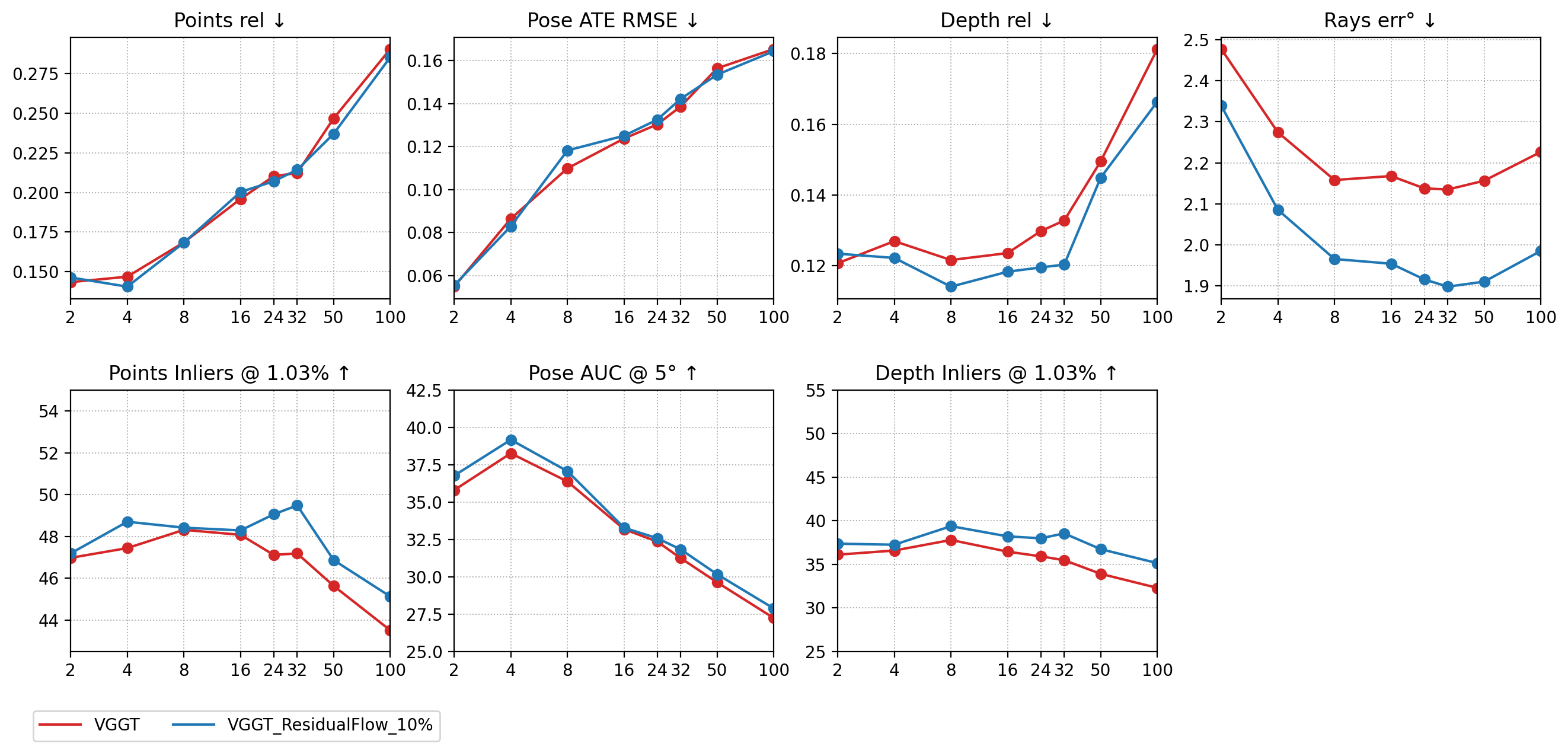}
\caption{Comparison of VGGT performance under default fine-tuning settings versus fine-tuning with a 10\% effective dataset size augmented by residual flow injection.}
\label{fig:vggt_comparison}
\end{figure}

We conduct an additional ablation study on dataset scale to demonstrate the effectiveness of our approach in addressing data scarcity during fine-tuning. We modulate the effective dataset size by varying the total number of scenes sampled from each dataset per epoch. To ensure a consistent number of training steps for a fair comparison, the number of epochs is adjusted inversely to the sampling rate; for instance, when reducing the number of scenes to 10\%, the total training duration is increased from 10 to 100 epochs. Figure~\ref{fig:pi3_ablation} illustrates a consistent performance improvement as the effective dataset size increases during $\pi^3$ fine-tuning. However, as shown in Figure~\ref{fig:pi3_comparison}, introducing residual flow injection with only 10\% of the training data yields higher benchmarking performance than increasing the dataset size tenfold. Similar results for VGGT are shown in Figure~\ref{fig:vggt_comparison}.

\begin{figure}[t]
\centering
\includegraphics[width=1\textwidth]{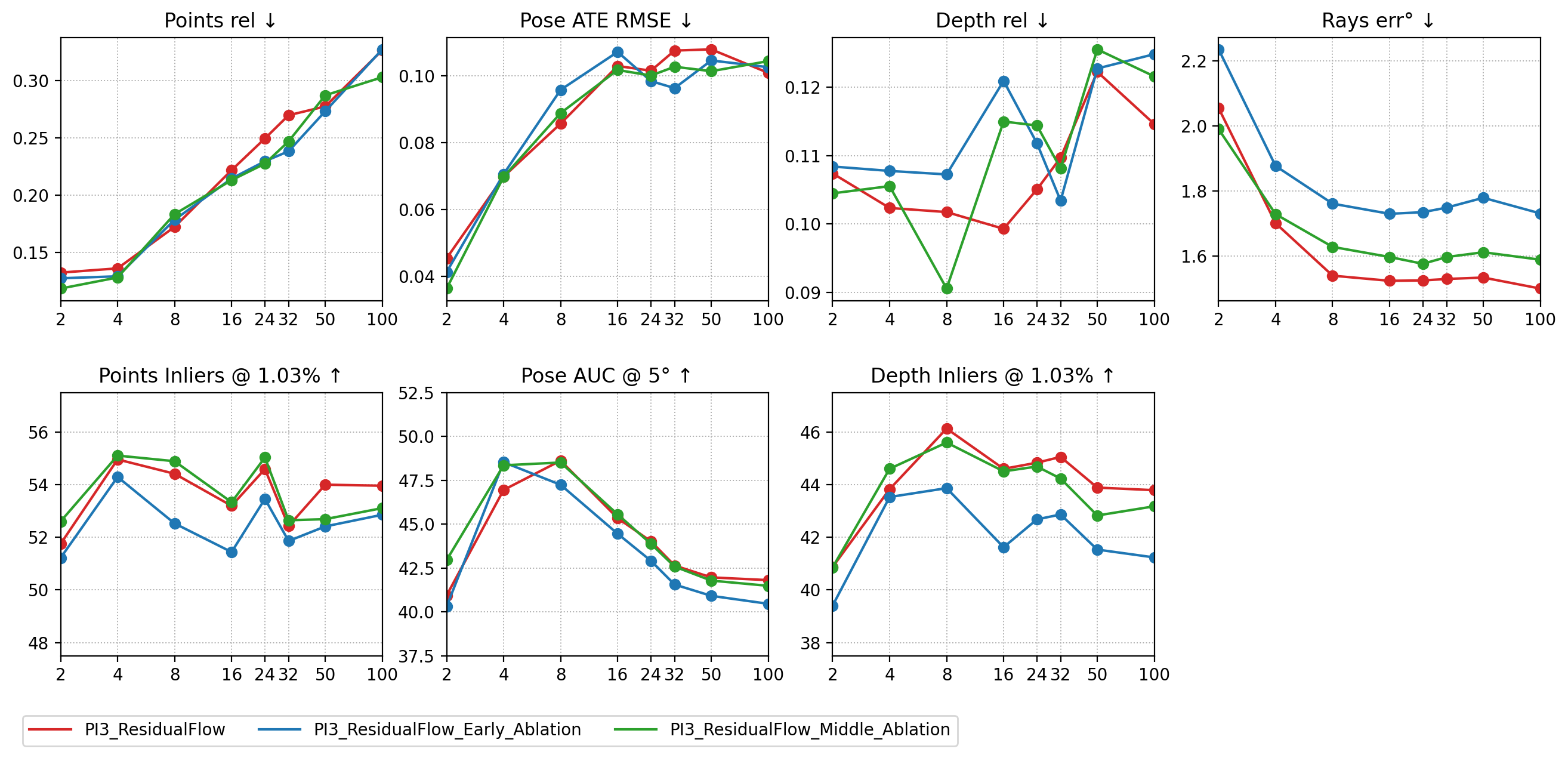}
\caption{Ablation study of residual flow injection locations within the $\pi^3$ model. Performance of default injection point is evaluated against injection occurring both prior to and within the middle layers of the aggregator module.}
\label{fig:pi3of_ablation}
\end{figure}

\begin{figure}[t]
\centering
\includegraphics[width=1\textwidth]{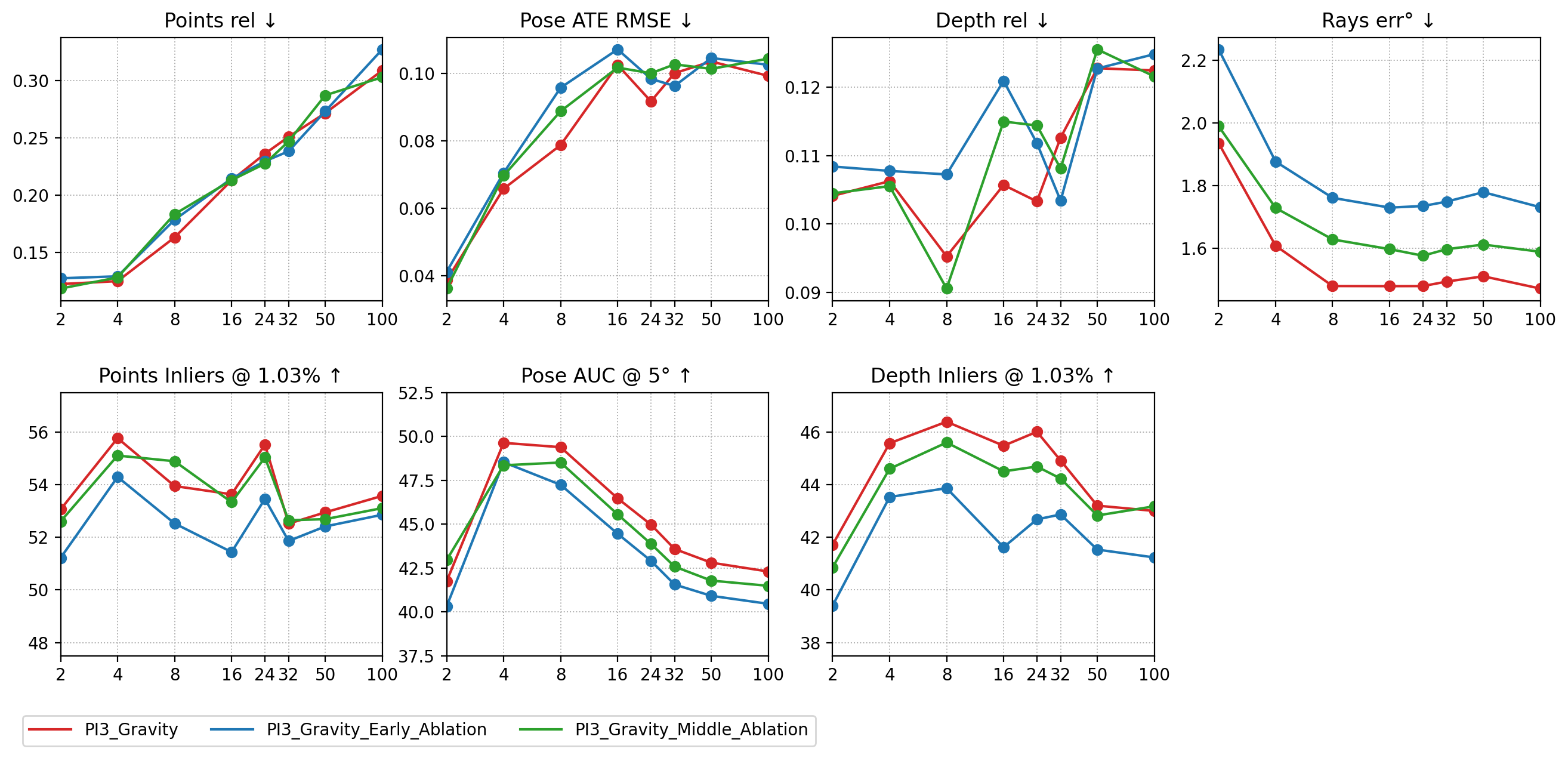}
\caption{Ablation study of gravity injection locations within the $\pi^3$ model. Performance of default injection point is evaluated against injection occurring both prior to and within the middle layers of the aggregator module.}
\label{fig:pi3rav_ablation}
\end{figure}

\begin{figure}[t]
\centering
\includegraphics[width=1\textwidth]{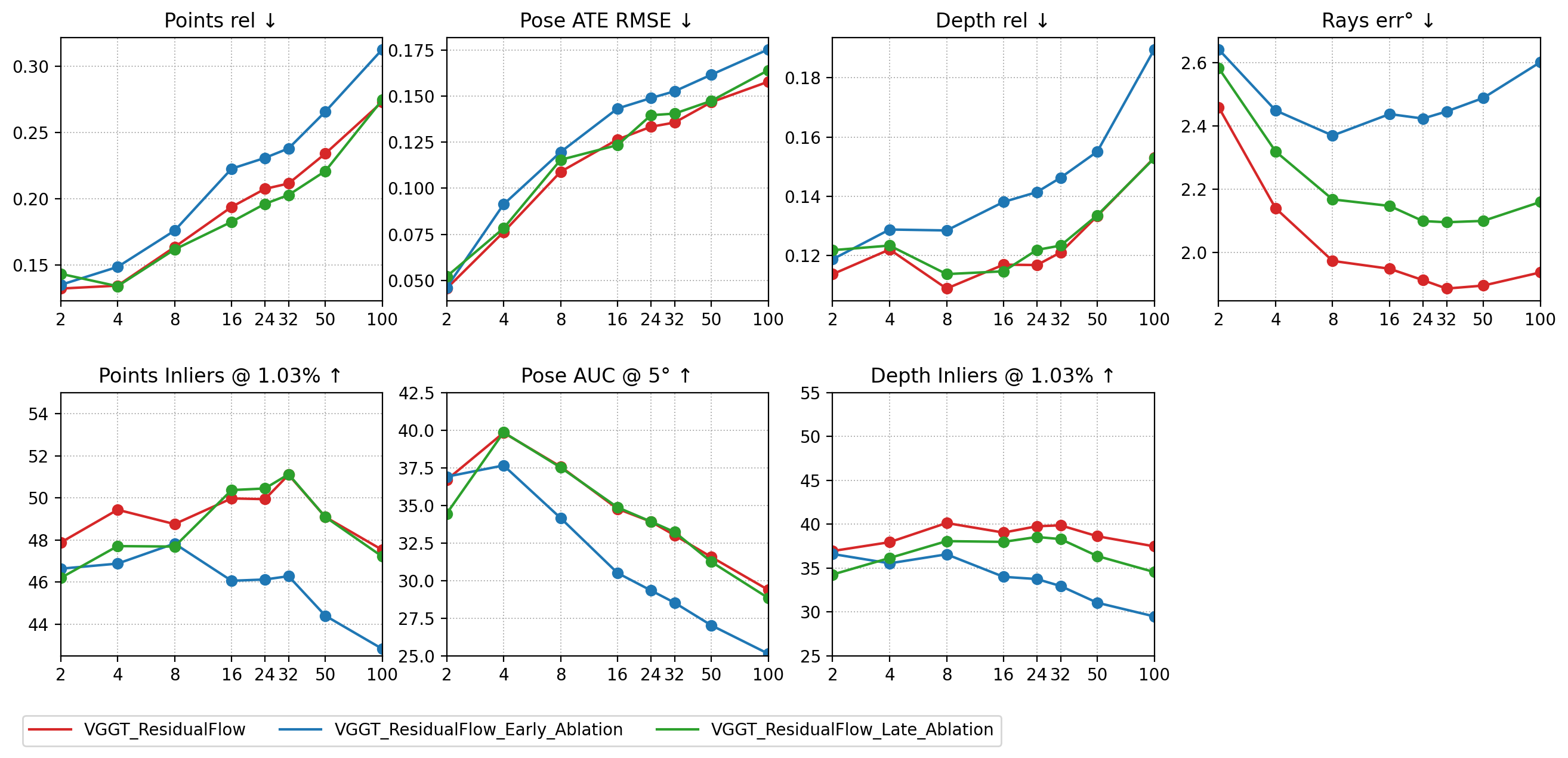}
\caption{Ablation study of residual flow injection locations within the VGGT model. Performance of default injection point is evaluated against injection occurring both before and after the aggregator module.}
\label{fig:vggtof_ablation}
\end{figure}

\begin{figure}[t]
\centering
\includegraphics[width=1\textwidth]{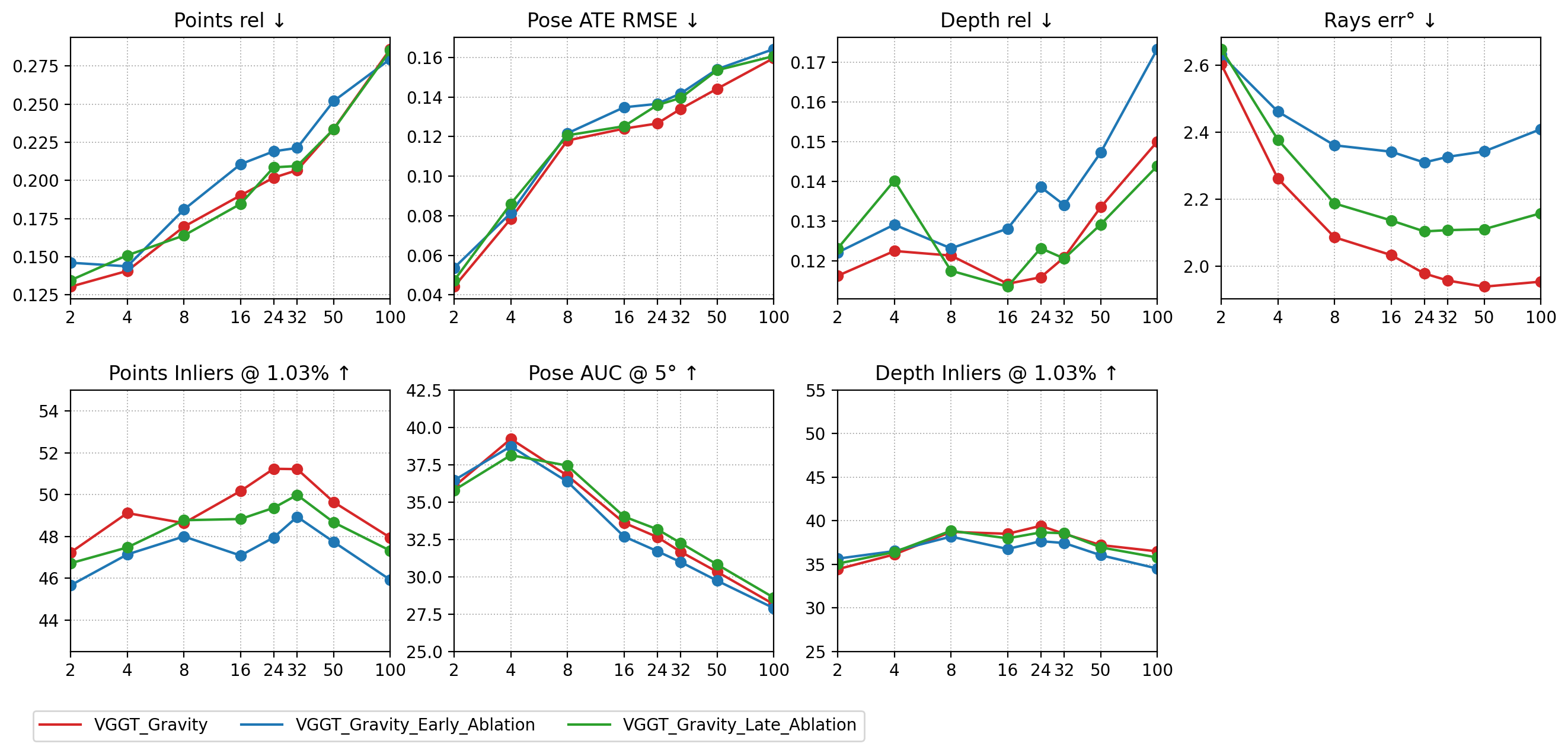}
\caption{Ablation study of gravity injection locations within the VGGT model. Performance of the default injection point is evaluated against injection occurring both before and after the aggregator module.}
\label{fig:vggtrav_ablation}
\end{figure}

We also provide comprehensive benchmarking results across varying view counts for our previously discussed injection location ablation in Figures~\ref{fig:pi3of_ablation}, ~\ref{fig:pi3rav_ablation}, ~\ref{fig:vggtof_ablation}, and ~\ref{fig:vggtrav_ablation}.

\clearpage

\FloatBarrier

\end{document}